\documentclass{article}

\usepackage[nonatbib,final]{nips_2016}

\usepackage[utf8]{inputenc} 
\usepackage[T1]{fontenc}    
\usepackage{hyperref}       
\usepackage{url}            
\usepackage{booktabs}       
\usepackage{amsfonts}       
\usepackage{amsmath}       
\usepackage{nicefrac}       
\usepackage{microtype}      

\usepackage[square,numbers]{natbib}
\usepackage{tikz}
\usetikzlibrary{positioning}
\usetikzlibrary{shapes}
\usetikzlibrary{shapes.geometric}

\title{Deep neural heart rate variability analysis}

\author{
  Tamas Madl \\
  Austrian Research Institute for Artificial Intelligence\\
  University of Manchester\\
  \texttt{tamas.madl@gmail.com} \\
}

\begin{document}

\maketitle

\begin{abstract}
  Despite of the pain and limited accuracy of blood tests for early recognition of cardiovascular disease, they dominate risk screening and triage. On the other hand, heart rate variability is non-invasive and cheap, but not considered accurate enough for clinical practice. Here, we tackle heart beat interval based classification with deep learning. We introduce an end to end differentiable hybrid architecture, consisting of a layer of biological neuron models of cardiac dynamics (modified FitzHugh Nagumo neurons) and several layers of a standard feed-forward neural network. The proposed model is evaluated on ECGs from 474 stable at-risk (coronary artery disease) patients, and 1172 chest pain patients of an emergency department. We show that it can significantly outperform models based on traditional heart rate variability predictors, as well as approaching or in some cases outperforming clinical blood tests, based only on 60 seconds of inter-beat intervals.
\end{abstract}

\section{Introduction}


Mammalian heart beats are induced by a control mechanism which generates electrical impulses to precipitate muscular contraction. Impulses are generated by the sinoatrial node (SAN), often called the heart's natural pacemaker, and are propagated through the atrioventricular node (AVN) and His-Purkinje system (PKJ), which are normally synchronized with the activity of SAN. A large body of previous work has investigated oscillator-based models of heart beat dynamics \cite{glass2001synchronization,di1998model,gois2009analysis} since the proposal of the Van der Pol oscillator \cite{vanderpol28}, or the more recent modified FitzHugh-Nagumo model of cardiac dynamics \cite{rogers1994collocation}. 

Heart beat dynamics are not only of academic interest, as time series of heart beat intervals (RR intervals) have been shown to be prognostic of several types of cardiovascular disease, and thus could be a basis for risk markers that are easy to measure non-invasively. Various families of disease predictors have been proposed in heart rate variability (HRV) analysis literature, including statistical measures such as the standard deviation of RR intervals, fractal and dynamical systems properties, entropy measures \cite{mansier1996linear,acharya2006heart,xhyheri2012heart}, symbolic dynamics, or, more recently, graph-based measures \cite{madl2016cinc}. Some models used artificial neural networks to combine and classify such predictors \cite{tsipouras2004automatic,acharya2004classification}. However, there is a large gap between these superficial predictors of HRV and the above-mentioned oscillator models exhibiting biologically plausible dynamics; and the practical application of the latter models to screening and prognosis has been neglected in literature.

Here, we introduce a biologically inspired, end-to-end differentiable neural network architecture based on the idea of non-linear coupling to simulated pacemaker neurons. Instead of tuning oscillator models to produce biologically plausible signals, we formulate the problem as finding the optimal ensemble of oscillators (and an associated non-linear classifier) yielding the smallest misclassification error when classifying heart beat time series into healthy or pathological.
We evaluate on data from stable but at-risk patients with vessel narrowing (coronary artery disease), and on chest pain (heart attack) patients.

In addition to significantly outperforming known HRV predictors on detecting patients at risk (screening), we also show that our model achieves a sensitivity and specificity comparable to cardiac Troponin I, the gold standard biomarker for detecting myocardial infarction \cite{hasic2003cardiac}, in a triage setting. The model only needs a 1 minute ECG segment, and does not require the >15-71min needed for point-of-care lab tests for excluding acute myocardial infarcts \cite{mccord2001ninety}. These preliminary results from 1645 patients suggest that deep neural HRV analysis may be useful both for risk screening (health check-ups) for prevention, and for triage (prioritization in emergency departments).

\section{Methods}

We start with the modified FitzHugh-Nagumo model, a biological neuron model proposed to account for cardiac impulse propagation \cite{rogers1994collocation} and argued to be able to simulate empirical data, including normal heart activity, ischemic conditions  \cite{berenfeld1996simulation}, atrial fibrillation, and other observed phenomena \cite{li2014three}.

The FitzHugh Nagumo equations \cite{Izhikevich2006FHN}, modified to account for cardiac activity \cite{rogers1994collocation}, can be written as follows (differences to standard parametrization - tied to biological parameters - are not important in our case, since the parameters are learned in a data-driven fashion):

\begin{equation}
    \dot{v_i} = v_i - v_i^3/3 - p_{i,1} w_i v_i + I
    \label{eq:fhn1}
\end{equation}
\begin{equation}
    \dot{w_i} = p_{i,2} (v_i-p_{i,3} w_i)
    \label{eq:fhn2}
\end{equation}

Here, $v_i$ corresponds to a membrane potential of neuron $i$, $w_i$ to its recovery variable, $I$ is an external input, and $p_{i,j}$ are model parameters, which (together with the input) govern the oscillatory dynamics.

We now assume that a population of modified FitzHugh-Nagumo (FHN) neurons is driven by an upstream source heart rate regulation (see Fig. 1 in \cite{voss2009methods}), by a stream of `action potentials' which constitute the model input. Our goal is to classify this input into healthy (normal sinus rhythm) or pathological (due to coronary artery disease / ischemia, i.e. insufficient blood flow due to e.g. narrowed arteries). To this end, we use a feed-forward neural network (NN), on top of the biologically inspired MFHN layer, to optimally fit and classify normal and pathological cardiac dynamics.

The NN receives a scalar firing rate from each FHN neuron as its input, assuming that FHN neurons `fire' when $v_i$ exceeds a firing threshold parametrized by $p_{i,4}$:

\begin{equation}
    f_i = \sum_{t=0}^{T} \begin{cases}
        v_i(t),& \text{if } v_i(t) \geq p_{i,4} \\
        0,              & \text{otherwise}
    \end{cases}
    \label{eq:input}
\end{equation}

The FHN neuron firing rates $f_i$ are fed into a feed-forward artificial neural network \cite{lecun2015deep} with hyperbolic tangent activation functions, several hidden layers, and a softmax output layer with two neurons to indicate ischemia (coronary artery disease patients) vs. normal heart beat dyanmics (healthy patients) -see Fig. \ref{fig:net}. Crucially, since all numerical computations - including both FHN dynamics, unrolled through time, and the feed-forward NN - are composed of a finite set of operations with known derivatives, the entire network is end-to-end differentiable. We use reverse-mode automatic differentiation (a generalization of backpropagation) with autograd \cite{maclaurinautograd} to train the network. 

More specifically, given known classes $y$, heart beat time series $rr$, and the parameter vector $P$ (which includes NN weights and biases as well as FHN neuron parameters), we use autograd to obtain the derivative of an L2 regularized cross-entropy objective function:

\begin{equation}
    J(P) = -\frac{\lambda}{n} ||P||^2 + \frac{1}{n} \sum_{k=1}^n y_k  \log O_I(rr_k, P) + (1-y_k)  \log (1-O_I(rr_k, P)),
    \label{eq:objective}
\end{equation}

where the first term represents L2 regularization, and $O_I(rr_k, P)$ represents normalized probabilities of time series $rr_k$ containing indications for ischemia, based on the described model. Having obtained the gradient of eq. \label{eq:objective} with autograd \cite{maclaurinautograd}, we use a stochastic optimizer with momentum, adam \cite{kingma2014adam}, to find the optimal model parameters. As the model is computationally expensive, hyperparameters such as number of layers, neurons per layer, minibatch size and $\lambda$ were adjusted by global black-box optimization (DIRECT \cite{gablonsky2001locally}) instead of a grid search to save computation time.

Initial NN weights were drawn from a normal distribution with $\sigma=0.1$ to break symmetry. Initial FHN neuron parameters $p_{i,1-4}$ were pre-optimized, also using DIRECT (due to the non-linearities in equations \eqref{eq:fhn1} and \eqref{eq:fhn2}, small changes in parameters can lead to very different dynamics; therefore random initialization of the FHN neurons seldom leads to good results). To avoid FHN parameters adapting to the random initial NN, they are clamped in the first $10\%$ epochs of the training procedure, and subsequently allowed to change such that the rough parameters found by DIRECT can be fine-tuned by gradient-based optimization. Training was stopped after 200 epochs or at convergence.

\tikzset{%
  mfhn neuron/.style={
    rectangle,
    draw,
    minimum size=1cm
  },
  every neuron/.style={
    circle,
    draw,
    minimum size=1cm
  },
  neuron missing/.style={
    draw=none, 
    scale=4,
    text height=0.333cm,
    execute at begin node=\color{black}$\vdots$
  },
}

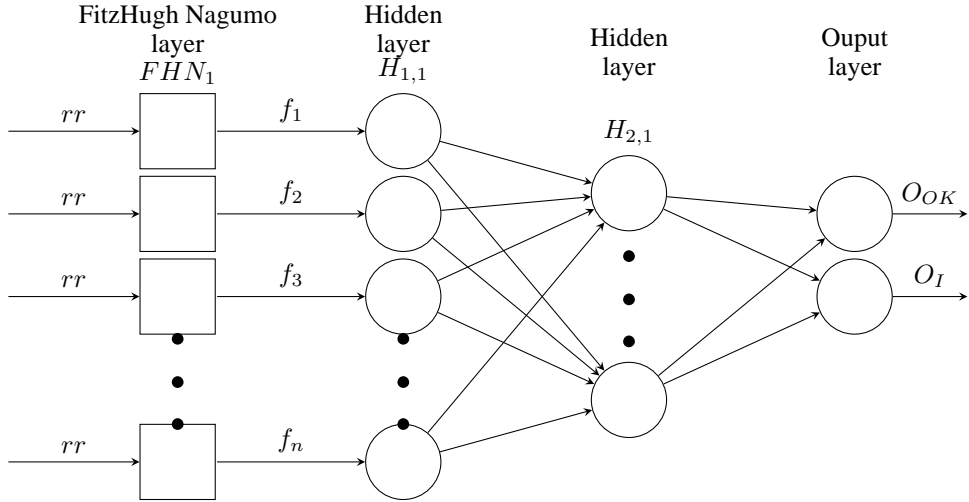
\begin{figure}
    \centering
\begin{tikzpicture}[x=1.5cm, y=1.1cm, >=stealth]

\foreach \m/\l [count=\y] in {1,2,3,missing,4}
  \node [mfhn neuron/.try, neuron \m/.try] (input-\m) at (-2,2.5-\y) {};

\foreach \m/\l [count=\y] in {1,2,3,missing,4}
  \node [every neuron/.try, neuron \m/.try] (hidden0-\m) at (0,2.5-\y) {};

\foreach \m [count=\y] in {1,missing,2}
  \node [every neuron/.try, neuron \m/.try ] (hidden-\m) at (2,2-\y*1.25) {};

\foreach \m [count=\y] in {1,2}
  \node [every neuron/.try, neuron \m/.try ] (output-\m) at (4,1.5-\y) {};

\foreach \l [count=\i] in {1,2,3,n}
  \draw [<-] (input-\i) -- ++(-1.5,0)
    node [above, midway] {$rr$};

\foreach \l [count=\i] in {1,2,3,n}
  \draw [<-] (hidden0-\i) -- ++(-1.65,0)
    node [above, midway] {$f_\l$};

\foreach \l [count=\i] in {1}
  \node [above] at (input-\i.north) {$FHN_\l$};

\foreach \l [count=\i] in {1}
  \node [above] at (hidden0-\i.north) {$H_{1,\l}$};

\foreach \l [count=\i] in {1}
  \node [above] at (hidden-\i.north) {$H_{2,\l}$};

\foreach \l [count=\i] in {OK,I}
  \draw [->] (output-\i) -- ++(1,0)
    node [above, midway] {$O_{\l}$};

\foreach \i in {1,...,4}
  \foreach \j in {1,...,2}
    \draw [->] (hidden0-\i) -- (hidden-\j);

\foreach \i in {1,...,2}
  \foreach \j in {1,...,2}
    \draw [->] (hidden-\i) -- (output-\j);

\foreach \l [count=\x from 0] in {FitzHugh Nagumo \\ layer \\ , Hidden \\ layer \\ , Hidden \\ layer, Ouput \\ layer}
  \node [align=center, above] at (\x*2-2,2) {\l};

\end{tikzpicture}
    \caption{Network architecture for classifying input heart beat interval time series $rr$ into ischemic (e.g. CAD) or normal, using scalar firing rates $f_i$ of FHN neurons to represent cardiac dynamics}
    \label{fig:net}
\end{figure}

\section{Experiments}

We tested the performance of our model in two settings, based on heart beat intervals extracted from Holter ECG data from the Telemetric Holter ECG Warehouse. To evaluate accuracy in risk screening, we used ECGs from 203 healthy participants\footnote{http://thew-project.org/Database/E-HOL-03-0202-003.html}, and 271 coronary artery disease patients\footnote{http://thew-project.org/Database/E-HOL-03-0271-002.html} (stable, but with vessel narrowing and excercise-induced ischemia). For investigating suitability for triage, we used data from 1172 chest pain patients from an emergency department\footnote{http://thew-project.org/Database/E-HOL-12-1172-012.html}.

Recordings were several hours long. To facilitate rapid decision making, we randomly extracted one minute segments from each. To obtain inter-beat intervals despite the motion artefacts and noise due to these portable devices recording from moving patients, we applied the noise-robust beat detection by \cite{kim2016simple}, and skipped segments with 1) less than 64 unique values or 2) less than 30 or more than 180 apparent heart rate. 100 segments were extracted from each patient to increase training data. Results below were obtained with 10-fold cross-validation, after randomization, stratified by patient ID, such that no data point from patients in the test set could ever occur in the training set. In the results tables, we also evaluate the inclusion of traditional HRV predictors into the model, under ``deep+traditional HRV'' (here, in addition to the FHN neurons, HRV predictors are included in the first layer).

\subsection{Risk screening}

Screening for cardiovascular events is currently based on blood test based risk scores, which have bad attendance (being invasive), as well as bad sensitivity and specificity in predicting significant coronary artery disease and mortalities: ROC AUC scores for common risk scores are around 0.64-0.68 \cite{versteylen2011comparison}, and up to $81\%$ of cardiovascular deaths occur in patients with ostensible risk scores under $40\%$  \cite{brindle2005accuracy}. 

Our goal was testing the described model in a screening setting, by classifying 474 patients into coronary artery disease vs. healthy, based purely on 1 minute RR intervals from ECG. This data can be obtained non-invasively, faster, and cheaper than blood tests. Table \ref{tbl:cad} compares the results of the model with a state of the art classification proposed by \cite{jovic2011electrocardiogram}, consisting of a random forest classifier with 30 decision trees and 23 HRV features, including standard deviation, root mean squared standard deviation, pNN20, approximate entropy, HRV triangular index, spatial filling index, central tendency measure, and the correlation dimension of the 2D embedding. We closely follow the setup of \cite{jovic2011electrocardiogram}.

\begin{table}[t]
  \caption{CAD screening: performance (sensitivity, specificity and ROC AUC) of deep neural HRV and traditional HRV (23 features combined using a random forest with 30 trees, as suggested in \cite{jovic2011electrocardiogram})}
  \label{tbl:cad}
  \centering
  \begin{tabular}{lccc}
    \toprule
Metric  & Deep neural HRV & Deep + traditional HRV & Traditional HRV  \\ \midrule
Sensitivity & $0.671 \pm 0.011$ & $\textbf{0.786} \pm 0.007$  & $0.701 \pm 0.018$  \\ 
Specificity & $0.897 \pm 0.002$ & $\textbf{0.917} \pm 0.001$  & $0.763 \pm 0.015$  \\ 
ROC AUC & $0.845 \pm 0.002$ & $\textbf{0.899} \pm 0.004$ & $0.803 \pm 0.013$  \\ 
    \bottomrule
  \end{tabular}
\end{table}

%

%

%


\subsection{Triage}

Cardiac Troponin I (cTnI) measurement is the gold standard in diagnosing acute myocardial infarction (AMI) \cite{hasic2003cardiac}. Sensitivities and specificities at admission lie between 63-69\% and 78\%-91\% \cite{apple2009role,olatidoye1998prognostic} (although a more recent study claims 0.94 ROC AUC \cite{body2011rapid}). A follow-up 4-10 hours later can improve these scores \cite{apple2009role} (however, making acute heart attack patients wait for hours is suboptimal). Even at admission, the test requires >15-71min for point of care testing \cite{mccord2001ninety}.

Here, we evaluated the described model in a rapid triage setting, classifying 1172 patients presenting to an emergency department with chest pain complaints into acute myocardial infarction (96) vs. angina non-acute coronary syndrome, or non-cardiac (1076) based on 1 minute ECG extracts. Table \ref{tbl:ami} shows the results. For this dataset, in addition to the comparison with the state of the art HRV model of \cite{jovic2011electrocardiogram} as above, we can also compare with the performance of triage based on troponin (cTnI), since cTnI levels were reported in our dataset. We used a threshold of 0.6, following \cite{antman2000myocardial}.

\begin{table}[t]
  \caption{AMI triage: performance (sensitivity, specificity and ROC AUC) of deep neural HRV and traditional HRV (23 features combined using a random forest with 30 trees, as suggested in \cite{jovic2011electrocardiogram})}
  \label{tbl:ami}
  \centering
  \begin{tabular}{lccccc}
    \toprule
Metric  & Deep neural HRV & Deep + traditional HRV  & Traditional HRV & cTnI (blood test) \\ \midrule
Sensitivity & $0.901 \pm 0.003$ & $0.904 \pm 0.002$ & $0.765 \pm 0.008$ & $\textbf{0.963} \pm 0.018$ \\ 
Specificity & $0.637 \pm 0.075$ & $\textbf{0.872} \pm 0.032$ & $0.620 \pm 0.064$ & $0.691 \pm 0.176$ \\ 
ROC AUC & $0.646 \pm 0.008$ & $0.745 \pm 0.007$ & $0.602 \pm 0.011$ & $\textbf{0.881} \pm 0.051$ \\ 
    \bottomrule
  \end{tabular}
\end{table}

\section{Conclusion}

Deep learning has made breakthroughs in several fields \cite{lecun2015deep}. However, the application to HRV analysis has been neglected, despite of the importance of non-invasive diagnostic tools, both for risk estimation in primary prevention settings, and for triage in acute emergency situations with limited resources. We have argued that these tools, combined with bio-inspired models of cardiac dynamics, show potential to significantly improve traditional HRV analysis, potentially contesting the claim that ``\textit{the potential for HRV to be used widely in clinical practice remains to be established}'' \cite{xhyheri2012heart}.

\subsubsection*{Acknowledgments}

Data used for this research was provided by the Telemetric and Holter ECG Warehouse of the University of Rochester (THEW), NY. We thank David Madl and Mate Toth for helpful comments.

\small

\bibliographystyle{plainnat}
\bibliography{nips}

\end{document}